\documentclass[lettersize,journal]{IEEEtran}
\usepackage{amsmath,amsfonts}
\usepackage{amssymb}
\usepackage{mathtools}
\usepackage{algorithm}
\usepackage{algpseudocode}
\usepackage{array}
\usepackage[caption=false,font=normalsize,labelfont=sf,textfont=sf]{subfig}
\usepackage{textcomp}
\usepackage{stfloats}
\usepackage{url}
\usepackage{verbatim}
\usepackage{graphicx}
\usepackage{cite}

\begin{document}

\title{MLATC: Fast Hierarchical Topological Mapping from 3D LiDAR Point Clouds Based on Adaptive Resonance Theory}
\author{Ryosuke~Ofuchi,~\IEEEmembership{Student~Member,~IEEE,}%
        Yuichiro~Toda,~\IEEEmembership{Member,~IEEE,}%
        Naoki~Masuyama,~\IEEEmembership{Member,~IEEE,}%
        and~Takayuki~Matsuno,~\IEEEmembership{Member,~IEEE}%
\thanks{This work has been submitted to the IEEE for possible publication. Copyright may be transferred without notice, after which this version may no longer be accessible.}
\thanks{Manuscript received April 19, 2021; revised August 16, 2021.}}

\markboth{Journal of \LaTeX\ Class Files,~Vol.~14, No.~8, August~2021}%
{Shell \MakeLowercase{\textit{et al.}}: A Sample Article Using IEEEtran.cls for IEEE Journals}


\maketitle

\begin{abstract}
This paper addresses the problem of building global topological maps from 3D LiDAR point clouds for autonomous mobile robots operating in large-scale, dynamic, and unknown environments. Adaptive Resonance Theory-based Topological Clustering with Different Topologies (ATC-DT) builds global topological maps represented as graphs while mitigating catastrophic forgetting during sequential processing. However, its winner selection mechanism relies on an exhaustive nearest-neighbor search over all existing nodes, leading to scalability limitations as the map grows. To address this challenge, we propose a hierarchical extension called Multi-Layer ATC (MLATC). MLATC organizes nodes into a hierarchy, enabling the nearest-neighbor search to proceed from coarse to fine resolutions, thereby drastically reducing the number of distance evaluations per query. The number of layers is not fixed in advance. MLATC employs an adaptive layer addition mechanism that automatically deepens the hierarchy when lower layers become saturated, keeping the number of user-defined hyperparameters low. Simulation experiments on synthetic large-scale environments show that MLATC accelerates topological map building compared to the original ATC-DT and exhibits a sublinear, approximately logarithmic scaling of search time with respect to the number of nodes. Experiments on campus-scale real-world LiDAR datasets confirm that MLATC maintains a millisecond-level per-frame runtime and enables real-time global topological map building in large-scale environments, significantly outperforming the original ATC-DT in terms of computational efficiency.
\end{abstract}


\section{Introduction}
\IEEEPARstart{W}{ith} the rapid advancement of robotics and sensing technologies, autonomous mobile robots are increasingly being deployed in dynamic real-world environments, including manufacturing, logistics, and infrastructure inspection. To execute tasks in such environments, autonomous robots must perceive their surroundings in real time, dynamically understand the environmental structure, and continuously update their motion plans accordingly.

Recent progress in Simultaneous Localization and Mapping (SLAM), particularly methods utilizing 3D point clouds acquired by LiDAR or depth sensors, has enabled the building of high-resolution metric maps~\cite{shan2020liosam, xu2022fastlio2, lou2023lidarvision}. 
While such metric maps preserve precise geometric information of the environment, they are not necessarily optimized for extracting semantic spatial information useful for navigation, such as the distribution of obstacles and traversable areas. 
In unknown environments without prior maps, sequential point cloud streams must be dynamically converted into spatial representations suitable for navigation tasks~\cite{muravyev2024navtopo}.

Topological maps, which represent the environment as graphs composed of nodes and edges, have long been recognized as effective solutions for autonomous navigation~\cite{thrun1998metrictopo}. 
By abstracting spatial information, topological maps enable efficient path planning and spatial recognition even in complex environments. 
Recently, methods for dynamically extracting topological maps from 3D point clouds through unsupervised learning have also been proposed. 
In particular, the ART-based Topological Clustering with Different Topologies (ATC-DT)~\cite{toda2024atcdt}, grounded in Adaptive Resonance Theory (ART)~\cite{carpenter1987art, grossberg1987art}, enables the building of a global topological map directly from raw point clouds.

However, a critical limitation of ATC-DT is its reliance on an exhaustive nearest-neighbor search over all existing nodes for each input point. 
This process entails a computational complexity proportional to the number of nodes, rendering real-time processing difficult in large-scale environments. 
Furthermore, the topological map built by ATC-DT maintains a dynamic graph structure subject to the frequent addition and deletion of nodes and edges, where the neighborhood information obtained at each step serves as the basis for subsequent structural updates. 
Consequently, achieving a balance between high accuracy and processing efficiency in the nearest-neighbor search becomes a paramount challenge.

To address this scalability limitation, we propose Multi-Layer ATC (MLATC), a hierarchical extension of ATC-DT. 
MLATC organizes topological maps into multiple layers with varying spatial resolutions and leverages this hierarchy to constrain the search range at each layer. 
This approach drastically reduces the number of distance calculations required for nearest-neighbor searches, thereby achieving high computational efficiency. 
The number of layers in MLATC is not fixed in advance; instead, it employs an adaptive layer addition mechanism that automatically deepens the hierarchy when existing layers become saturated. 
This allows the structural depth to adapt to the spatial scale while keeping the number of user-defined parameters minimal. 
The main contributions of this study are as follows:

\begin{itemize}
\item MLATC introduces a hierarchical nearest-neighbor search framework that limits the candidate set. Under mild assumptions and in practical settings, this reduces the linear search complexity of ATC-DT to a sublinear, approximately logarithmic trend with respect to the number of nodes, yielding substantial speedups in large-scale environments.
\item An adaptive layer addition mechanism automatically adjusts the hierarchy depth to the observed data, enabling the building of multi-layered topological maps while keeping the number of user-defined hyperparameters compact.
\item Experiments on synthetic and real-world large-scale point cloud datasets demonstrate that MLATC significantly improves computational efficiency while preserving the essential connectivity and topological characteristics of the maps built by the original ATC-DT.
\end{itemize}

This paper is organized as follows. 
Section~II summarizes existing methods related to topological map building. 
Section~III provides a detailed explanation of the proposed method, MLATC. 
Section~IV presents experimental results on both synthetic large-scale environments and real-world LiDAR datasets, including hierarchical map visualization and scalability analysis. 
Finally, Section~V concludes the paper and discusses future challenges and perspectives.

\section{Related Work}

Efficient and abstract representation of environmental structure is a fundamental requirement for autonomous navigation. 
A topological map is represented as a graph composed of nodes and edges. 
Nodes correspond to characteristic locations or regions in the environment, and edges represent spatial or traversable relationships between them. 
Compared to dense metric maps such as point clouds or meshes, this graph-based abstraction is lightweight and scalable, allowing compact environment representations suitable for long-term mapping and efficient path planning~\cite{kuipers2000spatial,thrun2005probabilistic}. 
In this work, we focus on topological maps whose nodes are defined as coordinates in Euclidean space and are built directly from LiDAR measurements. 
We investigate mechanisms for building such maps in a flexible and dynamically updated manner.

A well-established approach derives navigation-oriented topology from metric maps by extracting sparse connectivity structures. 
Many methods first build dense occupancy grids or distance fields online and then extract sparse free-space graphs on top of them. 
Sparse 3D Topological Graphs and related approaches build generalized Voronoi diagrams or skeletons over Euclidean signed distance fields (ESDFs), approximating environmental topology with a small number of nodes and edges to enable fast global planning~\cite{oleynikova2018sparse3d, chen2022fast3d}. 
Traversal Risk Graphs (TRG) further derive graph structures from elevation maps and attach traversability risk to edges to support safety-aware path planning~\cite{lee2025trg}. 
In these frameworks, however, the topological map is treated as a planning graph derived from dense metric representations. The graph structure itself is not directly self-organized in response to incoming point clouds, nor does it explicitly maintain multi-attribute node connectivity.

To handle differences in information content and spatial scale in a hierarchical manner, 3D Dynamic Scene Graphs (DSGs) represent places, rooms, buildings, objects, and agents as nodes and encode containment, adjacency, and spatio-temporal relations as edges, thereby unifying geometry and semantics in a single graph representation. 
Systems such as Kimera-DSG, Kimera-Multi, and Hydra build and optimize such DSGs online from sensor streams, using visual-inertial SLAM as the state-estimation backbone and dense metric-semantic reconstructions such as meshes and ESDFs as map representations, demonstrating high-level spatial perception and long-term navigation, including loop closure and multi-robot coordination~\cite{rosinol2021kimera,chang2021kimeramulti,hughes2022hydra}.
At the same time, these systems do not assume that LiDAR point clouds are directly learned as a lightweight topological representation, nor that the learned graph itself serves as a primary search structure in the LiDAR feature space.

Another stream of research learns topological maps directly from sequential observations through self-organizing neural networks. 
Representative models include the Self-Organizing Map (SOM)~\cite{kohonen1982som}, Growing Neural Gas (GNG)~\cite{fritzke1995gng}, and Growing When Required (GWR)~\cite{marsland2002gwr}, which incrementally adapt node positions and connectivity based on streaming data, forming graphs that reflect the geometric distribution of inputs. 
Hierarchical extensions such as the Growing Hierarchical SOM (GHSOM) and the Tree-Structured SOM (TS-SOM), as well as their spherical extensions (Spherical TS-SOM), grow map structures in a hierarchical or tree-shaped manner according to data distribution, enabling multi-resolution clustering and fast winner selection~\cite{dittenbach2000ghsom,koikkalainen1990tssom,koikkalainen1999tssom,yoshioka2022stssom}. 
In the GNG family, Multilayer Batch-Learning GNG (MBL-GNG) introduces batch learning and multi-layer organization to learn hierarchical topological structures ordered by abstraction level~\cite{toda2021mblgng}. 
For navigation, Growing Neural Gas with Different Topologies (GNG-DT)~\cite{toda2022gngdt,toda2023gngnavi} maintains independent topologies for attributes such as position, surface normal, and traversability, while selecting winner nodes only in position space. 
This approach obtains multiple attribute-wise clustering results online while preserving the underlying 3D spatial structure, and its effectiveness for spatial perception and navigation in unknown environments has been demonstrated. 
Nevertheless, as is common to GNG-based methods, structure updates through node insertion and deletion tend to cause the network size to grow with the scale of the environment, making it challenging to control memory usage and forgetting in long-term operation, and tuning numerous hyperparameters remains burdensome.

Adaptive Resonance Theory (ART)~\cite{carpenter1987art} provides a theoretical framework for balancing stability and plasticity in incremental learning. 
When the similarity between an input and an existing category representation exceeds a vigilance parameter, the associated node representation is updated; otherwise, a new node is generated as a separate cluster. 
Building on this principle, the Topological CIM-based ART (TCA) family employs the Correntropy-Induced Metric (CIM), an information-theoretic similarity measure robust to outliers, to form and adapt nodes and edges directly from continuous input streams~\cite{masuyama2019hcaea,masuyama2023paramfree}. 

Extending this foundation to global mapping for mobile robots, the ART-based Topological Clustering with Different Topologies (ATC-DT) algorithm was proposed~\cite{toda2024atcdt}. 
ATC-DT integrates the attribute-wise graph building of GNG-DT with the adaptive learning rule of ART, enabling sequential maintenance and expansion of multiple attribute layers such as position, surface normal, and reflectivity. 
This design mitigates catastrophic forgetting and improves memory efficiency, demonstrating the potential of ART-based topological learning for large-scale LiDAR mapping. 
However, winner selection in ATC-DT relies on exhaustive linear search over all existing nodes, causing the computational cost to grow linearly with network size and making it difficult to maintain real-time performance as the environment size increases.

Given this computational limitation, tree-based hierarchical space-partitioning structures are natural candidates for accelerating nearest-neighbor search in Euclidean space. 
Classical kd-trees~\cite{bentley1975kdtree} and octrees~\cite{meagher1982octree} achieve efficient queries for static or slowly changing point sets, but they are essentially static index structures and require costly rebuilding or rebalancing under frequent node insertions or deletions, which is incompatible with online learning. 
Incremental variants such as the ikd-tree~\cite{cai2021ikdtree} support sequential insertions, yet they typically assume that point locations remain fixed once inserted and do not accommodate the frequent node-position updates required by ART- or GNG-based learners such as ATC-DT.

In contrast, the proposed Multi-Layer ATC (MLATC) extends the ART-based learning framework into a hierarchical structure. 
By organizing topological maps across different resolutions, MLATC achieves fast winner selection and scalability while preserving the stability--plasticity balance of ART. 
In this sense, the proposed method bridges topological learning and hierarchical nearest-neighbor search, enabling scalable and multi-scale LiDAR-based topological mapping in large-scale environments.

\section{Method}
\subsection{Overview}
In the proposed method, the self-organizing framework of ATC-DT is extended by introducing a hierarchical structure and an adaptive layer addition mechanism. 
This enables faster nearest-neighbor search and the flexible building of topological maps that are scalable to large environments. 
This hierarchical extension is referred to as Multi-Layer ATC (MLATC).

ATC-DT is based on the learning rule of ART and dynamically organizes spatial structures while sequentially processing input points. 
In this study, this sequential learning mechanism is applied recursively, where nodes generated in lower layers serve as inputs for learning in the immediate upper layer. 
Through this process, hierarchical maps are built that capture spatial structures at different levels of resolution.

Each node explicitly records its parent--child relationship by referencing the lower-layer node from which it was derived during its generation process. 
This ensures not only the integrity of the topological structure within each layer but also consistent inter-layer references through node correspondences. 
By repeating this process, the topological map is progressively abstracted from local structures to global representations, and hierarchical referencing further improves search efficiency.

By utilizing the hierarchical structure built in this manner, nearest-neighbor search can be executed in a top-down fashion, iteratively narrowing the candidate set from the upper layers. 
Compared to an exhaustive search, this substantially reduces processing costs in large-scale environments. 
Moreover, while the conventional ATC-DT determines the abstraction level of the topological map solely through a single vigilance parameter, MLATC allows multiple granular spatial representations to be preserved simultaneously, thereby relaxing this constraint.

These characteristics are expected to be effective for applications such as semantic clustering based on path planning and connectivity. 
However, this paper does not address such applications and instead focuses primarily on accelerating nearest-neighbor search.

The hierarchical structure of MLATC is realized by a bottom-up approach that generates upper layers sequentially from the lowest layer, allowing dynamic adaptation to continuously incoming point cloud streams. 
On the other hand, since additional computational resources are required to build and maintain the upper-layer maps, the design of vigilance parameters from the second layer onward becomes a critical factor in balancing computational efficiency and memory usage. 
The influence of these design parameters on the overall processing cost of the algorithm is discussed in detail in Section~\ref{seq:vigi_para}, along with the respective components.

The definitions of the main variables used in this paper are listed in Table~\ref{tab:notation_proposed}.

\begin{table}[tb]
    \centering
    \caption{Notations in MLATC}
    \label{tab:notation_proposed}
    \begin{tabular}{l | p{6cm}}
        \hline
        \textbf{Notation} & \textbf{Definition} \\
        \hline
        $\rho_{\mathrm{vigi}}$ & Base vigilance parameter for the first layer \\
        $\ell$ & Layer index ($\ell \in \{1,2,\dots,L\}$) \\
        $\alpha$ & Vigilance parameter scaling factor \\
        $\rho_{\mathrm{search}}^{(\ell)}$ & Search threshold at layer $\ell$ \\
        $\mathcal{P},\, \mathbf{p}$ & Input point cloud and a point in $\mathcal{P}$ \\
        $\lambda$ & Number of training iterations per frame \\
        $\mathcal{W}^{(\ell)}$ & Set of candidate winner nodes at layer $\ell$, sorted by distance \\
        $G^{(\ell)}$ & Topological map at layer $\ell$ \\
        $\mathbf{h}_i^{(\ell),o}$ & Node attribute of type $o \in \{\mathrm{pos}, \mathrm{nor}\}$ at node $i$ in layer $\ell$ \\
        $h_i^{(\ell),\mathrm{tra}}$ & Traversability label of node $i$ at layer $\ell$ \\
        $\mathcal{C}_i^{(\ell)}$ & Set of child nodes of node $i$ at layer $\ell$ \\
        $m_i$ & Win count of node $i$ \\
        $g_{i,j}$ & Edge age between nodes $i$ and $j$ \\
        $g_{\mathrm{max}}$ & Edge age threshold for deletion \\
        \hline
    \end{tabular}
\end{table}

\subsection{ATC-DT Algorithm}
The learning algorithm of MLATC builds upon the framework of ATC-DT. 
Therefore, this section briefly introduces its baseline algorithm. 
ATC-DT~\cite{toda2024atcdt} builds independent topological maps for each attribute, such as position, normal vector, and traversability.
In MLATC, the lowest level of the topological map shares the same topological structure as ATC-DT. 
The set of topological maps $G$ is defined as
\[
G = (G^\mathrm{pos}, G^\mathrm{nor}, G^\mathrm{tra}).
\]
These maps share a common set of nodes $\mathcal{V} = \{1, 2, \dots, N\}$, where $N$ represents the number of nodes. 
Each component $G^\mathrm{o}$ for $o \in \{\mathrm{pos}, \mathrm{nor}\}$ is defined by a set of reference vectors $\{ \mathbf{h}_i^\mathrm{o} \}_{i \in \mathcal{V}}$ ($\mathbf{h}_i^\mathrm{o} \in \mathbb{R}^3$) and an edge set $\mathcal{E}^\mathrm{o} \subseteq \{\{i,j\} \subseteq \mathcal{V}\}$ as
\begin{equation}
    G^\mathrm{o} \coloneqq (\mathcal{V}, \{ \mathbf{h}_i^\mathrm{o} \}_{i \in \mathcal{V}}, \mathcal{E}^\mathrm{o}).
    \label{eq:G_posnor}
\end{equation}
Similarly, $G^\mathrm{tra}$ is defined with traversability labels $h_i^\mathrm{tra} \in \{0,1\}$ as attributes.
\begin{equation}
    G^\mathrm{tra} \coloneqq (\mathcal{V}, \{ h_i^\mathrm{tra} \}_{i \in \mathcal{V}}, \mathcal{E}^\mathrm{tra}).
    \label{eq:G_tra}
\end{equation}

The edge set $\mathcal{E}^\mathrm{pos}$ is built based on spatial neighborhoods in position space. 
In contrast, $\mathcal{E}^\mathrm{nor}$ and $\mathcal{E}^\mathrm{tra}$ are selected from the candidate set $\mathcal{E}^\mathrm{pos}$ according to the similarity of normal vectors and traversability, respectively. 
Consequently, the following inclusions hold.
\begin{equation}
    \mathcal{E}^\mathrm{nor} \subseteq \mathcal{E}^\mathrm{pos}, \quad \mathcal{E}^\mathrm{tra} \subseteq \mathcal{E}^\mathrm{pos}.
\end{equation}

The input data set $\mathcal{P}$ denotes a 3D point cloud of one frame acquired by a LiDAR sensor, where the number of points varies across frames. 
In the proposed method, for each frame, $\lambda$ points $\mathbf{p} \in \mathbb{R}^3$ are randomly sampled from $\mathcal{P}$ and used for learning. 
This procedure (i) absorbs differences in the number of points between frames to stabilize the processing time, and (ii) mitigates node drift caused by local biases in the input distribution.

At each step, the Euclidean distance $d_i$ between the input point $\mathbf{p}$ and the node's position vector $\mathbf{h}_i^\mathrm{pos}$ is calculated as
\begin{equation}
    d_i \coloneqq \bigl\lVert \mathbf{p} - \mathbf{h}^\mathrm{pos}_i \bigr\rVert_2.
\label{eq:euc_distance}
\end{equation}
Based on the distance $d_i$, the nearest node $s_1$ and the second-nearest node $s_2$ are selected as the first and second winner nodes, respectively. 
The processing for the input $\mathbf{p}$ is classified into the following three cases based on the distances to $s_1, s_2$ and the vigilance parameter $\rho_{\mathrm{vigi}} > 0$.

\begin{itemize}
\item[(a)] \textbf{Addition of a New Node} ($d_{s_1} > \rho_{\mathrm{vigi}}$)\\
If the input point is determined not to belong to the neighborhood of any existing node, a new node $N \gets |\mathcal{V}| + 1$ is added.
\begin{equation}
    \mathcal{V} \gets \mathcal{V} \cup \{N\}, \quad \mathbf{h}_N^\mathrm{pos} \gets \mathbf{p}.
    \label{eq:node_addition}
\end{equation}

\item[(b)] \textbf{Update of Existing Node} ($d_{s_1} \leq \rho_{\mathrm{vigi}} < d_{s_2}$)\\
The first winner node $s_1$ is updated by
\begin{equation}
    \mathbf{h}_{s_1}^\mathrm{pos} \leftarrow \mathbf{h}_{s_1}^\mathrm{pos} + \frac{1}{10m_{s_1}}(\mathbf{p} - \mathbf{h}_{s_1}^\mathrm{pos}),
    \label{eq:winner_update}
\end{equation}
where $m_i$ represents the number of times the $i$-th node has been selected as the first winner.
Additionally, for every neighbor $k$ of $s_1$, the update is performed with an attenuated learning rate.
\begin{equation}
    \mathbf{h}_k^\mathrm{pos} \leftarrow \mathbf{h}_k^\mathrm{pos} + \frac{1}{100m_k}(\mathbf{p} - \mathbf{h}_k^\mathrm{pos}).
    \label{eq:neighbor_update}
\end{equation}

\item[(c)] \textbf{Addition of Edge Connection} ($d_{s_2} \leq \rho_{\mathrm{vigi}}$)\\ 
In addition to the node update in (b), an edge is connected between $s_1$ and $s_2$, and its age is initialized.
\begin{equation}
    \mathcal{E}^\mathrm{pos} \gets \mathcal{E}^\mathrm{pos} \cup \{\{s_1, s_2\}\}.
    \label{eq:edge_reset}
\end{equation}
\end{itemize}

In cases (b) and (c), the age $g_{s_1,j}$ of each edge connected to $s_1$ is incremented, and edges exceeding the threshold $g_{\mathrm{max}}$ are removed.
\begin{equation}
    \mathcal{E}^\mathrm{pos} \gets \mathcal{E}^\mathrm{pos} \setminus \{\{i,j\} \mid g_{i,j} > g_{\mathrm{max}}\}.
    \label{eq:edge_removal}
\end{equation}
The threshold $g_{\mathrm{max}}$ is dynamically defined using the current set of edge ages $\Gamma$ and the set of deleted ages $\Gamma_{\mathrm{del}}$ as
\begin{equation}
    g_{\mathrm{max}} \coloneqq \overline{g}_{\mathrm{del}} \cdot \frac{|\Gamma_{\mathrm{del}}|}{|\Gamma_{\mathrm{del}}| + |\Gamma|} + g_{\mathrm{thr}} \cdot \left(1 - \frac{|\Gamma_{\mathrm{del}}|}{|\Gamma_{\mathrm{del}}| + |\Gamma|} \right),
    \label{eq:gmax}
\end{equation}
where $g_{\mathrm{thr}}$ is determined by the third quartile $\Gamma_{0.75}$ and the interquartile range $\mathrm{IQR}(\Gamma)$ of $\Gamma$.
\begin{equation}
    g_{\mathrm{thr}} \coloneqq \Gamma_{0.75} + \mathrm{IQR}(\Gamma).
    \label{eq:gthr}
\end{equation}

Thus, the topological map is adaptively updated for the input data. 
Based on the structure derived from this positional information, maps of attribute information such as normal vectors and traversability labels are also built. 
For detailed implementation and application examples, please refer to~\cite{toda2024atcdt}.

The overall flow of the algorithm is divided mainly into the winner selection process for the input vector $\mathbf{p}$ and the subsequent map update process, as summarized in Algorithm~\ref{alg:adaptive_map_update_preprocess} and Algorithm~\ref{alg:update_by_winner_nodes}.

\begin{algorithm}[tb]
\caption{\textsc{ATC-DT Main Process}}
\label{alg:adaptive_map_update_preprocess}
\begin{algorithmic}[1]
    \Require Input point cloud $\mathcal{P}$, topological map $G= (G^\mathrm{pos}, G^\mathrm{nor}, G^\mathrm{tra})$
    \Ensure Updated map $G$
    
    \State Acquire new point cloud $\mathcal{P}$
    \For{each of $\lambda$ iterations}
        \State Randomly select $\mathbf{p} \in \mathcal{P}$
        \State Compute $d_i$ by Eq.~\eqref{eq:euc_distance} for all $i \in \mathcal{V}$
        \State $s_1 = \arg\min_{i\in \mathcal{V}} d_i$
        \State $s_2 = \arg\min_{i \in \mathcal{V} \setminus \{s_1\}} d_i$
        \If{no corresponding nodes}
            \State $d_{s_1}, d_{s_2} \gets +\infty$
        \EndIf
        \State \Call{UpdateByWinners}{$\mathbf{p}$, $G^\mathrm{pos}$, $s_1$, $s_2$}
        \State Update $G^\mathrm{nor}$ and $G^\mathrm{tra}$ based on $G^\mathrm{pos}$
    \EndFor
    \State \Return Map $G$
\end{algorithmic}
\end{algorithm}

\begin{algorithm}[tb]
\caption{\textsc{UpdateByWinners}}
\label{alg:update_by_winner_nodes}
\begin{algorithmic}[1]
    \Require Input point $\mathbf{p}$, positional map $G^\mathrm{pos}$, 
             first and second winners $s_1$, $s_2$ (with distances $d_{s_1}$, $d_{s_2}$)
    \Ensure Updated positional map $G^\mathrm{pos}$
    
    \If{$d_{s_1} > \rho_\mathrm{vigi}$}
        \State Add a new node to $G^\mathrm{pos}$ by Eq.~\eqref{eq:node_addition}
    \Else
        \State $m_{s_1} \gets m_{s_1} + 1$
        \State Update $\mathbf{h}_{s_1}^\mathrm{pos}$ by Eq.~\eqref{eq:winner_update}
        \If{$d_{s_2} < \rho_\mathrm{vigi}$}
            \State Connect $s_1$ and $s_2$ in $\mathcal{E}^\mathrm{pos}$ by Eq.~\eqref{eq:edge_reset}
            \State $g_{s_1,s_2} \gets 0$
        \EndIf
        \For{each $k$ such that $\{s_1, k\} \in \mathcal{E}^\mathrm{pos}$}
            \State Perform an attenuated update of $\mathbf{h}_k^\mathrm{pos}$ by Eq.~\eqref{eq:neighbor_update}
            \State $g_{s_1,k} \gets g_{s_1,k} + 1$
        \EndFor
        \State Remove old edges from $\mathcal{E}^\mathrm{pos}$ by Eq.~\eqref{eq:edge_removal}
        \State Update threshold $g_\mathrm{max}$ by Eq.~\eqref{eq:gmax}
    \EndIf
\end{algorithmic}
\end{algorithm}

The original ATC-DT requires an exhaustive search to determine the first and second winner nodes. 
As the number of nodes increases, the computational cost grows linearly, limiting its applicability to autonomous mobile robots that require real-time processing. 
To address this issue, this paper introduces an efficient hierarchical nearest-neighbor search algorithm that balances computational efficiency while maintaining the quality of the learned topological map.

\subsection{Multi-layer ATC (MLATC)}
The hierarchical structure regards the fine-resolution topological map as the first layer, and the layer index is represented by $\ell \in \{1, 2, \dots, L\}$. 
The set of topological maps over all layers $G$ is defined as
\begin{equation}
    G \coloneqq \left\{ G^{(\ell)} \mid \ell \in \{ 1, 2, \dots, L \} \right\}.
    \label{eq:hierarchical_G}
\end{equation}
The map $G^{(\ell)}$ is built as
\begin{equation}
    G^{(\ell)} =
    \begin{cases}
        \left( G^{(1), \mathrm{pos}}, G^{(1), \mathrm{nor}}, G^{(1), \mathrm{tra}} \right), & \text{if } \ell = 1 \\
        G^{(\ell), \mathrm{pos}}, & \text{otherwise}
    \end{cases}
    \label{eq:hierarchical_Gl}
\end{equation}
where the first layer corresponds to three attributes: position, normal vector, and traversability, while the subsequent layers build a map based only on positional information. 
The structure for each attribute $G^{(\ell), o}$ ($o \in \{\mathrm{pos}, \mathrm{nor}, \mathrm{tra}\}$) follows the definition of the original ATC-DT (Eqs.~\eqref{eq:G_posnor} and \eqref{eq:G_tra}).

The vigilance parameter of each layer determines the spatial node density; a larger value results in sparser node placement. 
The vigilance parameter $\rho^{(\ell)}_\mathrm{vigi}$ in layer $\ell$ is determined using the base value $\rho_\mathrm{vigi}$ and the inter-layer ratio $\alpha$ by
\begin{equation}
     \rho^{(\ell)}_\mathrm{vigi} \coloneqq \alpha^{\ell - 1} \rho_\mathrm{vigi},
     \label{eq:vigi_l}
\end{equation}
where $\alpha$ is a hyperparameter that serves as a scaling factor for node density, directly affecting processing speed and memory consumption. 
The optimal value is discussed in Section~\ref{seq:vigi_para} based on theoretical analysis.

In MLATC, parent--child relationships between layers are introduced in addition to the structure of ATC-DT. 
Specifically, each node $i \in \mathcal{V}^{(\ell)}$ in layer $\ell$ has a set of child nodes $\mathcal{C}_i^{(\ell)}$, which is defined as a non-overlapping subset of the lower-layer node set $\mathcal{V}^{(\ell-1)}$.
\begin{equation}
    \mathcal{V}^{(\ell - 1)} \coloneqq \bigcup_{i \in \mathcal{V}^{(\ell)}} \mathcal{C}_i^{(\ell)}, \quad \mathcal{C}_i^{(\ell)} \cap \mathcal{C}_j^{(\ell)} = \emptyset \quad (\forall i \neq j).
    \label{eq:parent_child}
\end{equation}
This non-overlapping property is naturally ensured by the design in which each upper-layer node retains only the group of lower-layer nodes associated with it at the time of its generation. 
Through this structure, topological maps with different levels of abstraction can be built while consistently managing inter-layer reference relationships, thereby enabling fast search processing.

\begin{figure}[tb]
  \centering
  \includegraphics[width=0.9\linewidth]{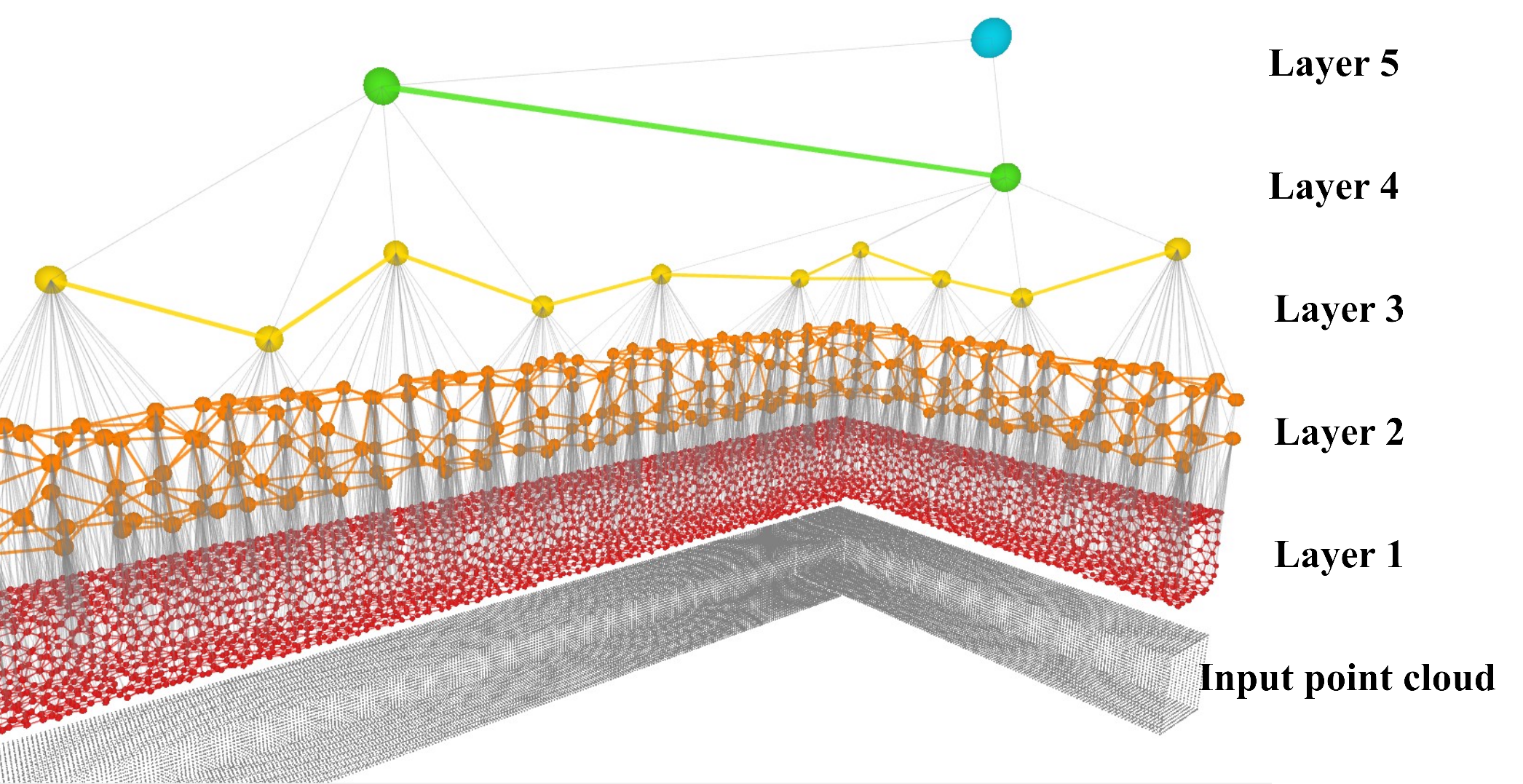}
  \caption{Hierarchical structure of the proposed MLATC. The first layer $G^{(1)}$ is a fine-resolution topological map directly learned from LiDAR point clouds. Upper layers $G^{(2)}, \dots, G^{(L)}$ summarize disjoint child node sets into coarser-resolution nodes, and the top layer maintains a single root node. Parent--child links between layers are used to constrain the search space during hierarchical nearest-neighbor search.}
  \label{fig:mlatc_hierarchy}
\end{figure}

Figure~\ref{fig:mlatc_hierarchy} illustrates the overall hierarchical structure of MLATC. 
Nodes in the first layer $G^{(1)}$ form a fine-resolution topological map directly learned from LiDAR point clouds, while upper layers $G^{(2)}, \dots, G^{(L)}$ recursively summarize non-overlapping child node sets into coarser-resolution nodes, and the top layer maintains a single root node.

The overall process flow of the proposed MLATC is shown in Algorithm~\ref{alg:hierarchical_atc_dt}. 
This algorithm consists of two phases: the exploration phase and the learning phase.

In the first phase, a search for neighboring nodes is conducted in each layer starting from the top layer, building a candidate set of winner nodes $\mathcal{W}^{(\ell)}$ for the input data. 
In the subsequent second phase, using the preserved $\mathcal{W}^{(\ell)}$, a learning process based on the winner nodes is applied sequentially from the bottom layer ($\ell = 1$). 
For each input point $\mathbf{p}$, the first and second winner nodes ($s_1$, $s_2$) are selected from the candidate set. 
If the distance between $\mathbf{p}$ and $s_1$ exceeds the vigilance parameter, a new node is added; otherwise, nodes and edges are updated. 
These update processes follow the learning algorithm (Algorithm~\ref{alg:update_by_winner_nodes}) of the original ATC-DT.

\begin{algorithm}[tb]
\caption{MLATC Main Process}
\label{alg:hierarchical_atc_dt}
\begin{algorithmic}[1]
    \Require Input point cloud $\mathcal{P}$, topological map $G$
    \Ensure Updated map $G$
    
    \State Acquire new point cloud $\mathcal{P}$
    \For{each of $\lambda$ iterations}
        \State Randomly select $\mathbf{p} \in \mathcal{P}$
        \State $\mathcal{W} \gets$ \Call{HierarchicalNNS}{$\mathbf{p}$, $G$}
        \State $\ell \gets 1$
        \While{$\ell \leq L$}
            \State Select $s_1, s_2$ with smallest $d^{(\ell)}$ in $\mathcal{W}^{(\ell)}$
            \State \Call{UpdateByWinners}{$\mathbf{p}, G^{(\ell),\mathrm{pos}}, s_1, s_2$}
            \If{new node added}
                \If{$\ell = L$ and $|\mathcal{V}^{(L)}| = 2$}
                    \State Add new layer $G^{(L+1)}$
                    \State $L \gets L+1$
                \EndIf
                \State $\ell \gets \ell + 1$
            \Else
                \State \textbf{break} 
            \EndIf
        \EndWhile
        \State Update $G^\mathrm{nor}$ and $G^\mathrm{tra}$ based on $G^\mathrm{pos}$
    \EndFor
    \State \Return Map $G$
\end{algorithmic}
\end{algorithm}

When a node is newly added, it is regarded as an input point for the upper layer, and similar learning processes are recursively applied to the upper layers. 
This recursive update builds a structure where learning propagates continuously from the lower layers to the upper layers. 
From the second layer onward, each upper-layer node records the first winner as its parent and the corresponding lower-layer input node as its child. 
Connections based on this parent--child structure enable efficient search from the upper layers, contributing to accelerated topological map building.

Additionally, when the number of nodes in the top layer reaches two, a new layer is added. 
This follows the design policy of always maintaining a single root node in the top layer. 
At the time of layer addition, the first node is inherited as a new node in the upper layer, and the second node serves as the input to the upper layer, allowing for a smooth transition of the learning process. 
As a result, the hierarchical structure can be dynamically expanded without predefining the spatial scale, while avoiding bottlenecks associated with the addition of new layers.

\subsection{Hierarchical Nearest-Neighbor Search}
\label{seq:NS}
In this algorithm, for an input point $\mathbf{p}$, the hierarchical topological map is traversed sequentially from the top layer ($\ell = L$) to the bottom layer, progressively narrowing down the candidate winner nodes at each layer. 
The candidate set is restricted by introducing a search threshold $\rho^{(\ell)}_{\mathrm{search}}$ defined as
\begin{equation}
    \rho^{(\ell)}_{\mathrm{search}} \coloneqq \sum_{i=1}^{\ell} \rho^{(i)}_{\mathrm{vigi}},
    \label{eq:search_radius}
\end{equation}
which is derived from the assumption of a worst-case configuration where a child node lies on the boundary of its parent node's vigilance region. 
Under this assumption, the maximum reachable distance within the vigilance regions across layers is upper-bounded by the cumulative sum of the vigilance parameters.

On the other hand, the node position update defined in Eq.~\eqref{eq:winner_update} may, in principle, allow a node to drift outside this threshold if the input remains biased in a single direction for an extended period. 
In practice, however, LiDAR point clouds inherently contain irregularities derived from real-world environments, and each frame provides input points by random sampling. 
Therefore, such sustained directional bias is unlikely to occur. 
Moreover, because winner nodes are updated incrementally within their local neighborhoods, nodes tend to remain in the high-density regions of the observed distribution. 
From these considerations, the proposed search threshold functions as a conservative upper bound.

At the top layer, the map maintains a single root node ($|\mathcal{V}^{(L)}| = 1$). 
Since filtering based on distance is not performed at this stage, the candidate set $\mathcal{W}^{(L)}$ is initialized as the entire set of nodes $\mathcal{V}^{(L)}$.
\begin{equation}
    \mathcal{W}^{(L)} = \mathcal{V}^{(L)}.
    \label{eq:winner_set_top}
\end{equation}

In layers other than the top layer ($\ell < L$), the set of candidate nodes $\mathcal{W}^{(\ell)}$ includes only those nodes $c$ from the child node set $\mathcal{C}_w$ belonging to the winner node $w \in \mathcal{W}^{(\ell+1)}$ of the immediate upper layer $\ell+1$, for which the Euclidean distance $d_c^{(\ell)}$ from the input point $\mathbf{p}$ is less than or equal to the threshold $\rho^{(\ell)}_{\mathrm{search}}$. 
This process is formulated as
\begin{equation}
    \mathcal{W}^{(\ell)} = \left\{ 
        c \in \bigcup_{w \in \mathcal{W}^{(\ell+1)}} \mathcal{C}_w^{(\ell+1)} 
        \,\middle|\, 
        d^{(\ell)}_{c} \le \rho^{(\ell)}_{\mathrm{search}} 
    \right\}.
    \label{eq:winner_set}
\end{equation}
Note that $\mathcal{C}_w^{(\ell+1)}$ denotes the set of child nodes in layer $\ell$ associated with the parent node $w$ in layer $\ell+1$.

By limiting candidates from parent nodes based on the layered structure and confining the search range in each layer to a local area, the search space can be significantly reduced compared to the linear search over all nodes in the original ATC-DT. 
This gradual narrowing allows the computational complexity to approach approximately logarithmic scaling relative to the number of nodes, enabling fast and scalable nearest-neighbor search.

\begin{algorithm}[tb]
\caption{\textsc{HierarchicalNNS}}
\label{alg:search}
\begin{algorithmic}[1]
    \Require Input point $\mathbf{p} \in \mathcal{P}$, topological map $G$
    \Ensure Winner sets $\mathcal{W} = \{\mathcal{W}^{(1)}, \dots, \mathcal{W}^{(L)}\}$
    \State Initialize $\mathcal{W}^{(\ell)} \gets \emptyset$ for all $\ell = 1, \dots, L$
    \For{$\ell = L$ \textbf{down to} $1$}
        \If{$\ell = L$}
            \State Set $\mathcal{W}^{(L)}$ by Eq.~\eqref{eq:winner_set_top}
        \Else
            \State Filter child nodes by Eq.~\eqref{eq:winner_set}
        \EndIf        
        \State Sort $\mathcal{W}^{(\ell)}$ in ascending order of $d^{(\ell)}_w \quad (w \in \mathcal{W}^{(\ell)})$
    \EndFor
    \State \Return Winner sets $\mathcal{W}$
\end{algorithmic}
\end{algorithm}

\subsection{Vigilance Parameter Design}
\label{seq:vigi_para}
The vigilance parameter is a factor that determines the resolution of the topological map, and it should be set to reflect the robot's embodiment and navigation requirements. 
In MLATC, the base vigilance parameter $\rho^{(1)}_{\mathrm{vigi}}$ is determined based on the robot specifications, as in the original ATC-DT.

The vigilance parameter $\rho^{(\ell)}_{\mathrm{vigi}}$ at the upper layers is directly related to search efficiency and memory usage. 
In MLATC, it is defined by Eq.~\eqref{eq:vigi_l} using the scaling factor $\alpha$. 
When $\alpha$ is large, the number of layers decreases while the number of child nodes per parent node increases, resulting in a larger branching factor. 
Conversely, when $\alpha$ is small, the layer structure becomes deeper, increasing the number of search steps. 
Therefore, $\alpha$ should be designed to minimize the overall search cost by considering this trade-off between branching factor and depth.

In the following analysis, we assume that point clouds are uniformly distributed on a 2D plane with a sufficiently large area $A$. 
This assumption is justified by the fact that LiDAR point clouds are locally distributed on 2D surfaces and, at a sufficiently large spatial scale, even complex terrains can be approximated globally by a plane. 
Let $N^{(\ell)}$ denote the number of nodes in layer $\ell$ and $\sigma^{(\ell)}$ be the node density per unit area. 
Then, the following relationship holds as
\begin{equation}
    \sigma^{(\ell)} = \frac{N^{(\ell)}}{A}.
    \label{eq:node_area}
\end{equation}
Since a new node is added when no existing node lies within the vigilance parameter, the node density is considered inversely proportional to the area of the vigilance circle. 
Introducing a proportionality constant $\kappa$ and substituting Eq.~\eqref{eq:vigi_l}, we obtain the approximation
\begin{equation}
    \sigma^{(\ell)} \approx \frac{\kappa}{\pi\left(\alpha^{\ell-1}\rho^{(1)}_{\mathrm{vigi}}\right)^2}.
    \label{eq:areal_density}
\end{equation}
Here, $\kappa$ is a dimensionless constant that reflects the deviation between the idealized coverage and the actual node placement. 
Based on the densest hexagonal circle packing in the plane, it can be regarded as at most $\kappa_{\max} = \pi/(2\sqrt{3})$.

Substituting this approximation into Eq.~\eqref{eq:node_area} and reorganizing with the inter-layer relationship of vigilance parameters (Eq.~\eqref{eq:vigi_l}), the node ratio between layers $\ell + 1$ and $\ell$ is derived as
\begin{equation}
    \frac{N^{(\ell)}}{N^{(\ell + 1)}} = \alpha^2.
    \label{eq:childset_alpha}
\end{equation}
By accumulating this ratio across layers and considering that MLATC always maintains a single root node in the top layer (i.e., $N^{(L)}=1$), the required number of layers $L(N, \alpha)$ for a given number of nodes at the bottom layer $N \coloneqq N^{(1)}$ is determined as
\begin{equation}
    L(N,\alpha) = \left\lceil \frac{\log N}{2\log\alpha} \right\rceil + 1.
    \label{eq:layer_function}
\end{equation}

Next, we analyze the number of candidates for the nearest-neighbor search in each layer. 
Since the search threshold follows Eq.~\eqref{eq:search_radius}, the candidate size can be estimated as the product of the circular area with radius $\rho^{(\ell)}_{\mathrm{search}}$ and the density from Eq.~\eqref{eq:areal_density}. 
Substituting the inter-layer relation (Eq.~\eqref{eq:vigi_l}) and rearranging, we obtain
\begin{equation}
    k^{(\ell)} \approx \kappa\left(\frac{1-\alpha^{-\ell}}{1-\alpha^{-1}}\right)^2,
    \label{eq:kl_alpha_re}
\end{equation}
where $k^{(\ell)}$ is uniformly bounded across layers by $\kappa \le k^{(\ell)} \le k_0(\alpha)$, with $k_0(\alpha)=\kappa\bigl(\alpha/(\alpha-1)\bigr)^2$.

The number of distance computations in each layer is approximately proportional to the size of the child sets maintained by the parent nodes selected in the previous layer. 
Using the relation in Eq.~\eqref{eq:childset_alpha}, the average size is estimated to be $\alpha^2\,k^{(\ell+1)}$. 
In addition, the cost of selecting the first and second winner nodes can be upper bounded by $O\!\left((k^{(\ell)})^2\right)$ through sequential comparisons. 
By introducing the positive constant coefficients $c_{\mathrm{d}}$ and $c_{\mathrm{s}}$, the overall search cost across all layers is approximated as
\begin{equation}
    T(N,\alpha)
    =\sum_{\ell=1}^{L(N,\alpha)-1}
      \left\{
        c_{\mathrm{d}}\,\alpha^2\,k^{(\ell+1)}(\alpha)
        + c_{\mathrm{s}}\,\bigl(k^{(\ell)}(\alpha)\bigr)^2
      \right\}.
    \label{eq:total_cost}
\end{equation}
From this expression, the minimization problem is formulated as follows. 
Let $r \coloneqq (c_{\mathrm{s}}\kappa)/c_{\mathrm{d}}$ be a dimensionless parameter representing the relative weight of the winner-selection cost to the distance-computation cost. 
The objective function $\widetilde{H}(\alpha;r)$ for $\alpha > 1$ is defined as
\begin{equation}
\widetilde{H}(\alpha;r)
=\frac{\alpha^2}{\log\alpha}\!\left(\frac{\alpha}{\alpha-1}\right)^{\!2}
+\frac{r}{\log\alpha}\!\left(\frac{\alpha}{\alpha-1}\right)^{\!4}.
\label{eq:H_simplified}
\end{equation}
The optimal scaling factor is then given by minimizing $\widetilde{H}(\alpha;r)$.

By sweeping $r\in[0,10]$ for Eq.~\eqref{eq:H_simplified}, we confirmed that the optimal value $\alpha^\ast(r)$ is uniquely determined and increases monotonically as $r$ grows. 
Representative results are summarized in Table~\ref{tab:alpha_star}.

\begin{table}[tb]
  \centering
  \caption{Representative optimal values $\alpha^\ast$ over $r$}
  \label{tab:alpha_star}
  \begin{tabular}{c c}
    \hline
    $r=(c_{\mathrm{s}}\kappa)/c_{\mathrm{d}}$ & $\alpha^\ast$ \\
    \hline
    0.0  & 2.891 \\
    0.5  & 3.246 \\
    1.5  & 3.665 \\
    5.0  & 4.470 \\
    10.0 & 5.157 \\
    \hline
  \end{tabular}
\end{table}

In practical applications, it is difficult to estimate $r$ precisely in advance. 
However, considering that even in the densest packing of sequentially added nodes the bound $\kappa \le \pi/(2\sqrt{3})$ holds, stable performance can be achieved by setting $\alpha$ in the range of approximately $3.5$--$4.5$, regardless of moderate variations in the relative balance between distance computations and sorting costs.

The space complexity is bounded above by a geometric series because the number of nodes decreases by a factor of $\alpha^{-2}$ at each layer according to Eq.~\eqref{eq:childset_alpha}. 
With $N$ denoting the number of nodes at the bottom layer, the complexity is expressed as
\[
\mathcal{O}\!\left(\frac{N}{1-\alpha^{-2}}\right).
\]
Thus, configurations with $\alpha \approx 1$ should be avoided because the space complexity diverges. 
On the other hand, increasing $\alpha$ monotonically reduces space complexity, while from the perspective of time complexity, the optimum lies around $\alpha = 3.5$--$4.5$. 
Therefore, a balanced design must carefully consider the trade-off between these two aspects.

\section{Experiments}
This section presents the experimental validation of the proposed Multi-Layer ATC (MLATC) in comparison with a non-hierarchical baseline. 
Our evaluation focuses on three aspects: (1) quantifying computational scalability with respect to the number of nodes, (2) examining the preservation of geometric and topological structure across hierarchical layers using qualitative visualization and simple statistics, and (3) demonstrating applicability to large-scale real-world mapping. 
For a fair comparison, the baseline corresponds to the original ATC-DT with the hierarchical extension disabled, using identical vigilance parameters and update rules. 
All implementations were executed in a single CPU thread. 
The experimental environment and parameters are summarized in Table~\ref{tab:machine_specs} and Table~\ref{tab:experiment_params}, respectively. 
Unless otherwise specified, the scaling factor is fixed at $\alpha = 4.0$.

\begin{table}[tb]
  \centering
  \caption{Machine specifications.}
  \label{tab:machine_specs}
  \begin{tabular}{ll}
    \hline
    \textbf{Item} & \textbf{Details} \\
    \hline
    CPU & Intel Core i7-1360P (12C/16T) \\
    OS & Ubuntu 22.04 LTS \\
    Compiler & GCC 11.4 / Clang 14.0 \\
    Build option & \texttt{-O3 -DNDEBUG} \\
    \hline
  \end{tabular}
\end{table}

\begin{table}[tb]
  \centering
  \caption{Common experimental parameters.}
  \label{tab:experiment_params}
  \begin{tabular}{ll}
    \hline
    \textbf{Parameter} & \textbf{Value} \\
    \hline
    Training iterations per frame $\lambda$ & 4000 \\
    Vigilance parameter $\rho_{\mathrm{vigi}}$ & $0.5\,\mathrm{m}$ \\
    Scaling factor $\alpha$ & 4.0 \\
    Square size (synthetic) & $20\,\mathrm{m}$ \\
    Frame translation & $10\,\mathrm{m}$ (fixed direction) \\
    \hline
  \end{tabular}
\end{table}

\subsection{Synthetic Data Evaluation}
To evaluate scalability under controlled conditions, we considered an unbounded synthetic stream of frames: at each frame, $\lambda$ points were uniformly sampled within a $20\,\mathrm{m}$ square window, and the window was translated by $10\,\mathrm{m}$ in a fixed direction. 
As frames accumulate, the total number of nodes $N$ increases monotonically.

\begin{figure}[tb]
  \centering
  \includegraphics[width=0.9\linewidth]{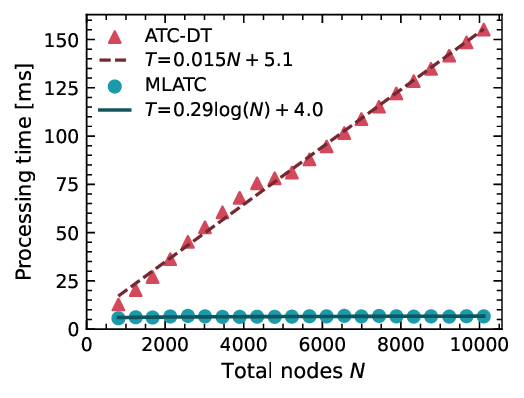}
  \caption{Processing time comparison on synthetic data ($N\!\le\!10^4$).}
  \label{fig:syn_efficiency}
\end{figure}

\begin{figure}[tb]
  \centering
  \includegraphics[width=0.9\linewidth]{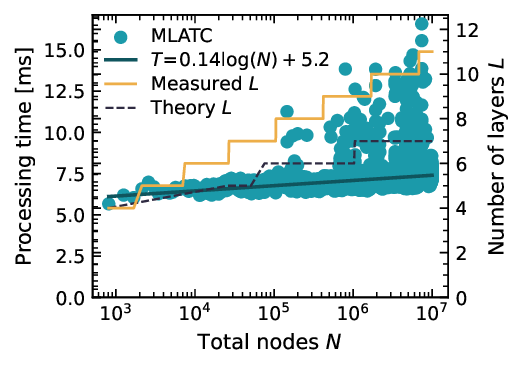}
  \caption{Scalability of MLATC on synthetic data up to $N=10^7$. The network formed up to 11 active layers during the sequence.}
  \label{fig:syn_scalability}
\end{figure}

For the comparison between MLATC and the non-hierarchical baseline, we analyzed only the initial part of this stream and recorded statistics while $N$ grew up to approximately $10^4$, corresponding to the first 22 frames (Fig.~\ref{fig:syn_efficiency}). 
To investigate larger scales, we then continued processing the same synthetic stream with MLATC only until the total number of nodes reached $N = 10^7$, and Fig.~\ref{fig:syn_scalability} reports the resulting scalability curve. 
In all cases, the reported runtime $T(N)$ includes nearest-neighbor search, node updates, and edge updates.

The baseline exhibited almost perfectly linear scaling ($T\!\approx\!0.015N\!+\!1.1$, $R^2\!=\!1.000$), whereas the proposed method showed a much slower, sublinear increase. 
A logarithmic regression model $T\!\approx\!0.29\log N\!+\!4.2$ yielded $R^2\!=\!0.50$, indicating that the runtime grows only weakly with $N$ and is partly dominated by frame-to-frame fluctuations. 
Beyond $N\!\approx\!3\times10^2$, MLATC consistently outperformed the baseline. 
At $N\!\approx\!10^4$, the proposed method achieved a mean per-frame time of $6.6$\,ms compared with $155.1$\,ms for the baseline, corresponding to a $24\times$ improvement in throughput. 

In the extended sequence ($N\!\le\!10^7$), the runtime of MLATC remained nearly constant at $7.3\!\pm\!0.4$\,ms, and a linear fit did not exhibit a clear dependence on $N$ ($R^2\!=\!0.09$). 
This near-constant behavior over several orders of magnitude in $N$ is empirically compatible with the at most logarithmic search complexity derived in Section~\ref{seq:vigi_para}, although within the available range it is difficult to distinguish conclusively between logarithmic and constant scaling. 
Across this sequence, $95\%$ of frames were processed within $7.7$\,ms.

\subsection{Real-World LiDAR Evaluation}
\begin{figure}[tb]
  \centering
  \includegraphics[width=0.9\linewidth]{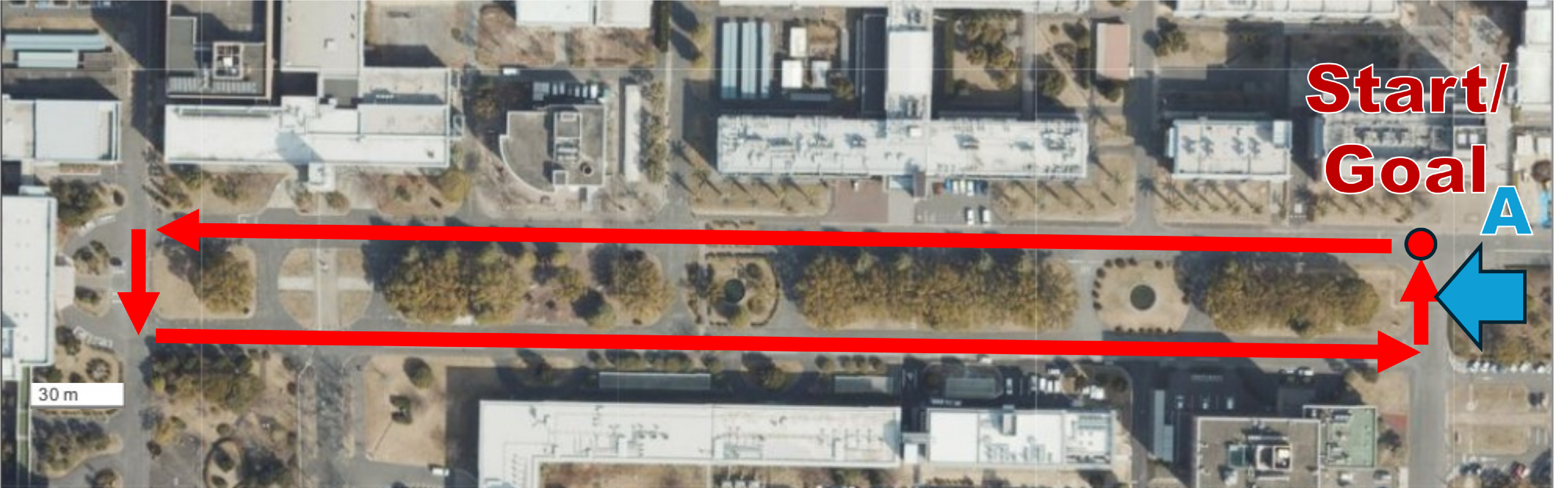}
  \caption{Aerial view of the real-world experiment. Red: LiDAR trajectory; blue arrow: reference point A.
  Basemap: aerial photograph by Geospatial Information Authority of Japan (GSI)~\cite{gsi2025}, edited to overlay the trajectory and marker.}
  \label{fig:overview_photo}
\end{figure}

For real-world validation, 3D point clouds were collected along an $800$\,m route on the Okayama University campus using a trolley-mounted LiDAR system equipped with a MID-360 sensor (Fig.~\ref{fig:overview_photo}). 
The trajectory included areas with dense vegetation, building facades, and open corridors, providing diverse geometric densities. 
Pose information acquired by FAST-LIO2~\cite{xu2022fastlio2} was used for global registration, and the resulting world-frame point clouds were directly processed by the clustering module. 
Both methods employed the same number of training iterations per frame $\lambda$ and vigilance parameter $\rho_{\mathrm{vigi}}$, differing only in whether the hierarchical extension was enabled (MLATC) or disabled (ATC-DT baseline).

In the small-scale regime ($N\!\le\!2\times10^4$), the proposed method achieved a mean processing time of $3.6$\,ms per frame, compared with $71.0$\,ms for the baseline ($19.7\times$ faster). 
Regression analysis yielded $T\!\approx\!0.80\log N-3.5$ ($R^2\!=\!0.68$) for MLATC and $T\!\approx\!0.006N+0.9$ ($R^2\!=\!0.998$) for the baseline, again indicating clear linear growth for ATC-DT and a much weaker, sublinear growth for MLATC despite increased variance due to non-uniform point densities along the route. 
On the full dataset ($N_{\max}=2.2\times10^5$), MLATC runtime averaged $4.5\!\pm\!0.8$\,ms, with the 90th and 95th percentiles at $5.7$\,ms and $6.0$\,ms, respectively, well below the $10$\,ms real-time threshold. 
Throughout the sequence, the hierarchy dynamically maintained up to six active layers, adapting to the spatial scale and density variations of the environment.

\begin{figure}[tb]
  \centering
  \includegraphics[width=0.9\linewidth]{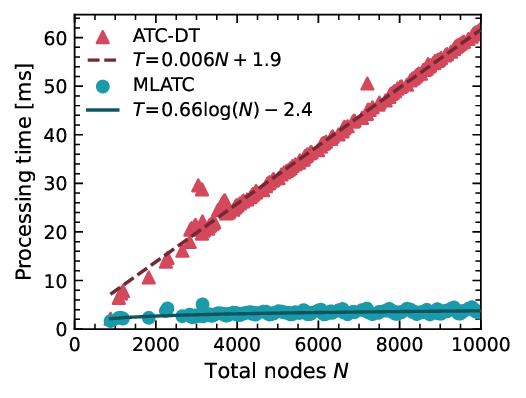}
  \caption{Per-frame processing time on real-world LiDAR data (small-scale regime).}
  \label{fig:real_efficiency}
\end{figure}

\begin{figure}[tb]
  \centering
  \includegraphics[width=0.9\linewidth]{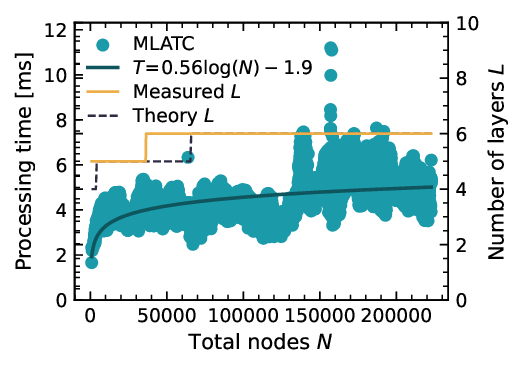}
  \caption{Scalability on real-world LiDAR data up to $N\!\approx\!2.2\times10^5$.}
  \label{fig:real_scalability}
\end{figure}

\subsection{Hierarchical Map Visualization}
\begin{figure}[tb]
  \centering
  \subfloat[Layer~1]{%
    \includegraphics[width=\linewidth]{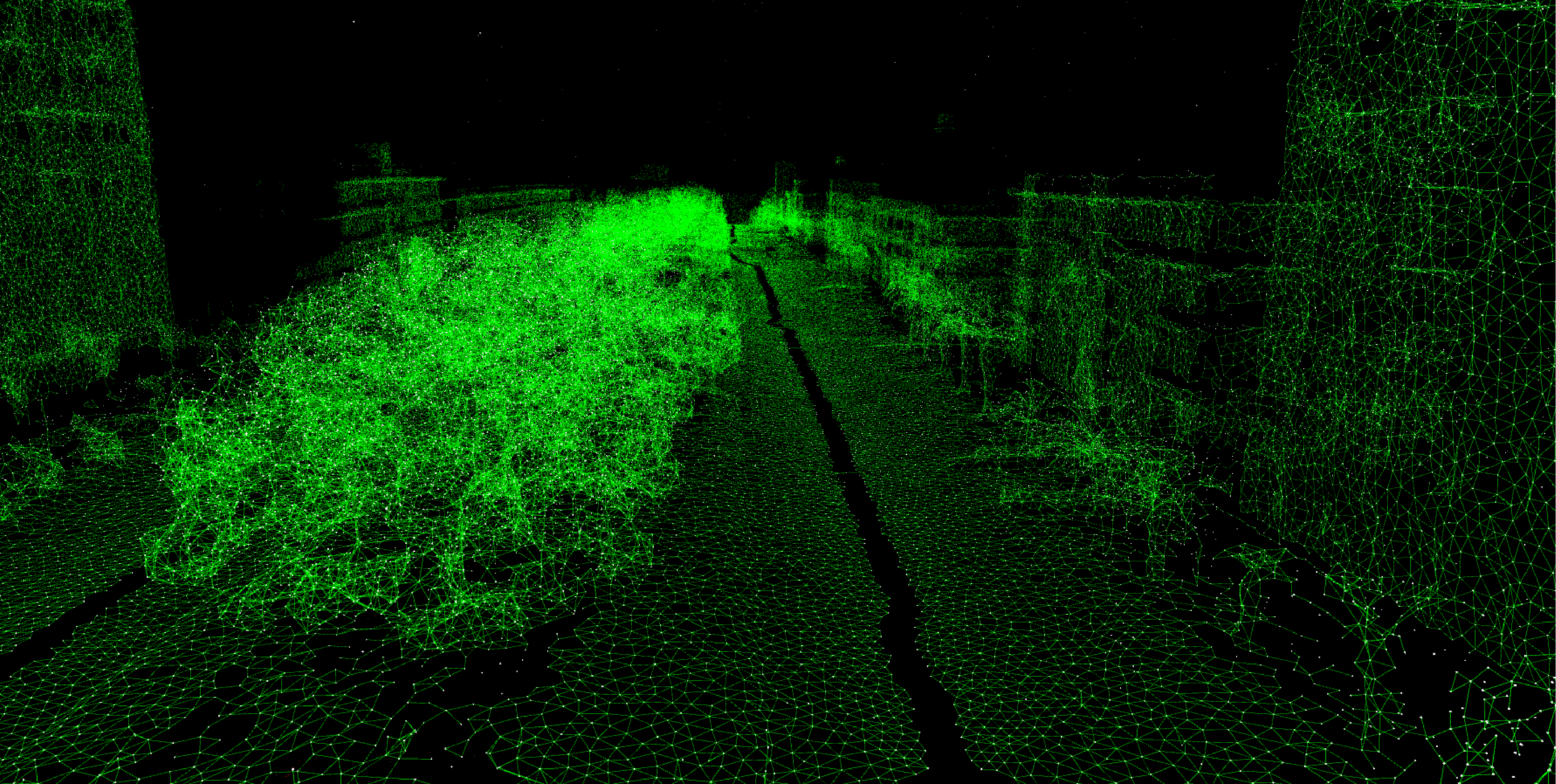}%
    \label{fig:layer1_map}}\\[0.3em]
  \subfloat[Layer~2]{%
    \includegraphics[width=\linewidth]{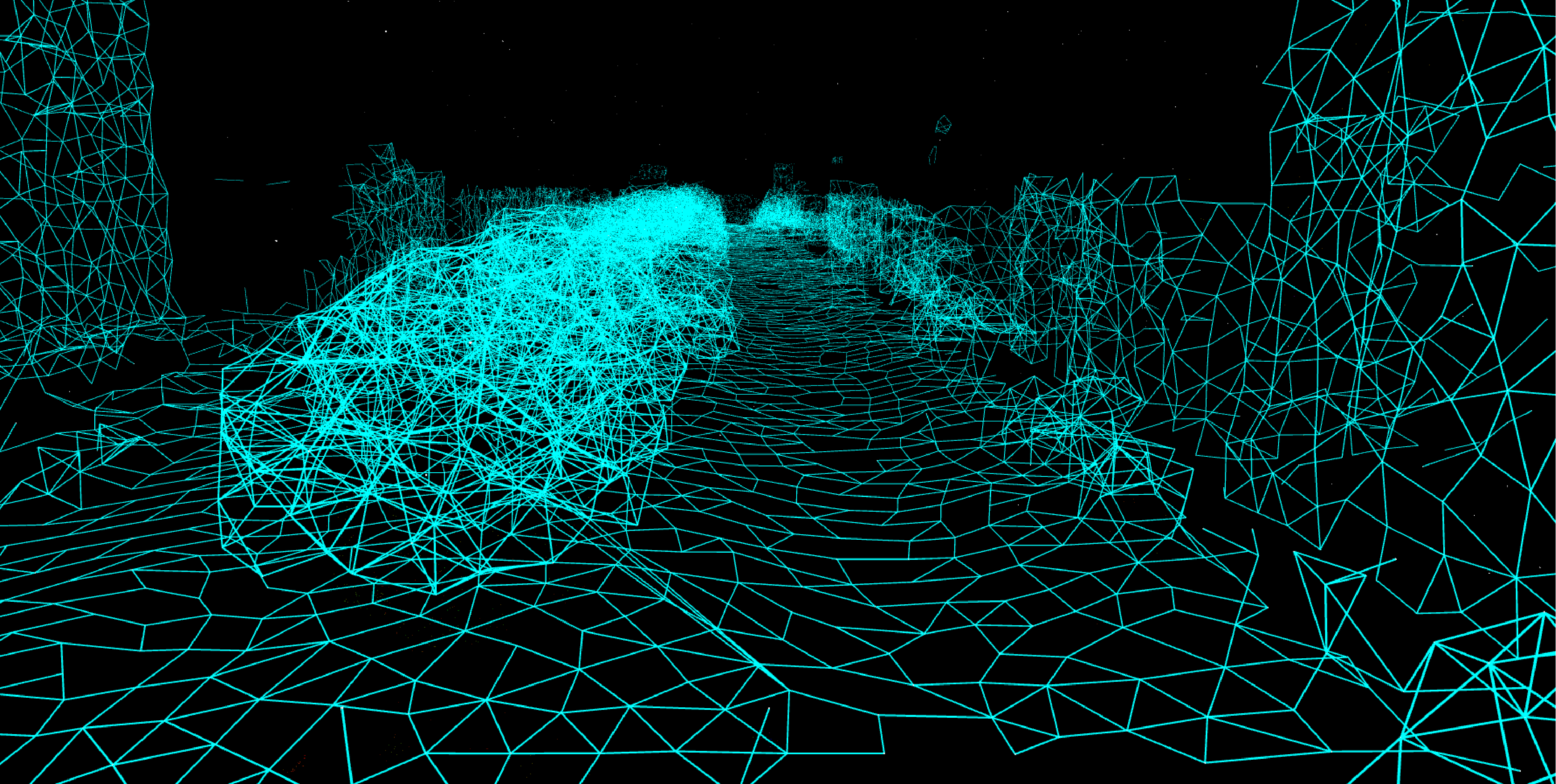}%
    \label{fig:layer2_map}}\\[0.3em]
  \subfloat[Layer~3]{%
    \includegraphics[width=\linewidth]{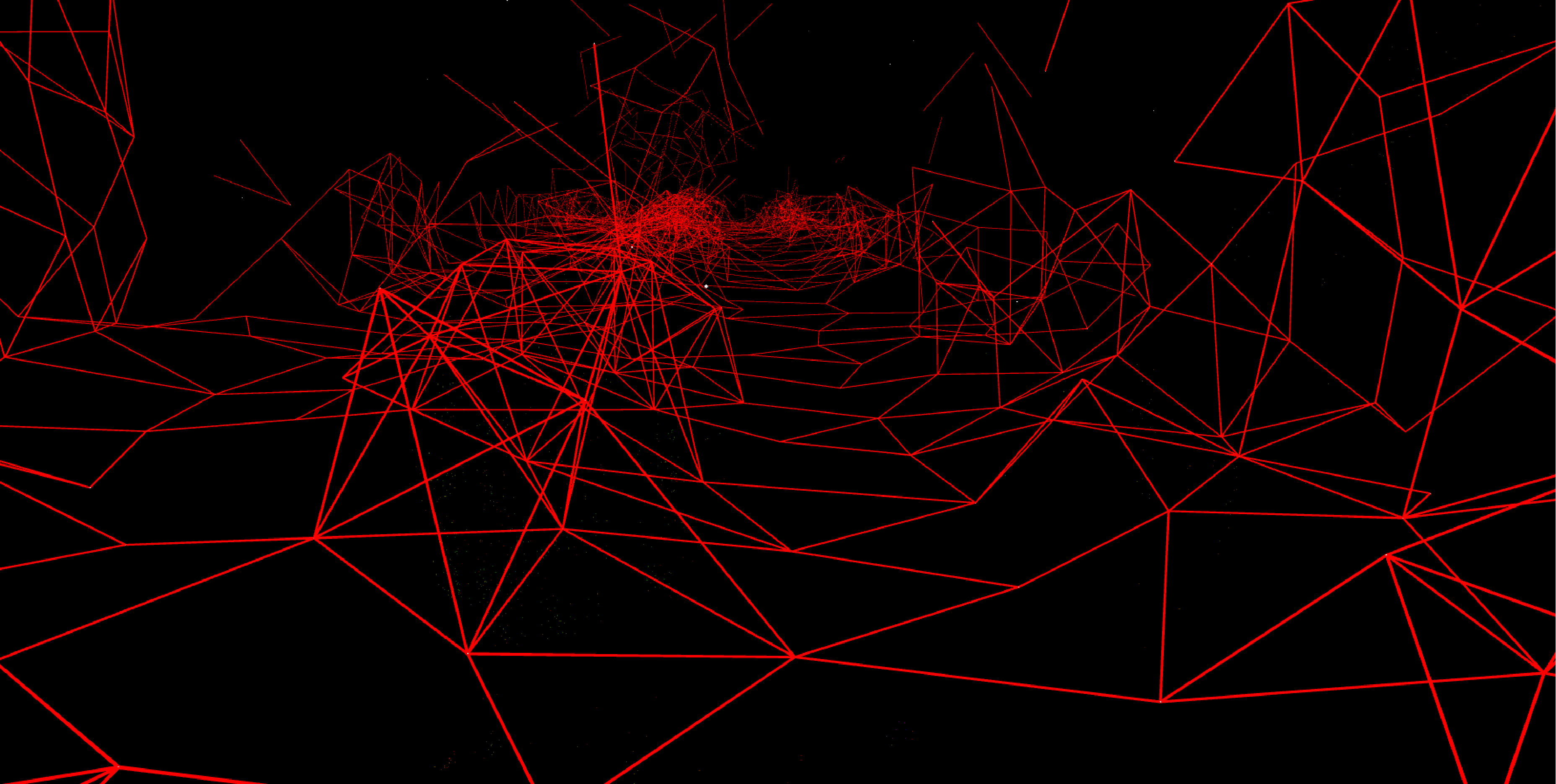}%
    \label{fig:layer3_map}}
  \caption{Hierarchical topological maps around reference point~A in Fig.~\ref{fig:overview_photo}. Nodes and edges of Layer~1, Layer~2, and Layer~3 are visualized from the viewpoint of A, illustrating how the same area is represented with progressively coarser spatial resolution as the layer index increases.}
  \label{fig:hierarchical_maps_real}
\end{figure}

Figures~\ref{fig:layer1_map}--\ref{fig:layer3_map} illustrate representative topological maps reconstructed around reference point~A in Fig.~\ref{fig:overview_photo}. 
Layer~1 represents fine-resolution local planar connectivity, Layer~2 merges spatially adjacent clusters into continuous surface-level structures, and Layer~3 provides coarse abstractions corresponding to corridor and junction levels. 
At the current stage, these layers mainly differ due to the change in vigilance parameters, which results in varying spatial densities but does not yet carry explicit semantic meaning. 
Future work will assign functional roles to different layers, such as navigation-level abstraction or semantic reasoning, to better exploit the multi-scale hierarchy.

\begin{table}[tb]
  \centering
  \caption{Layer-wise node and edge statistics}
  \label{tab:layer_stats}
  \begin{tabular}{lccc}
    \hline
    Layer & $|\mathcal{V}|$ & $|\mathcal{E}|$ & Node ratio [\%] \\
    \hline
    1 & 222{,}763 & 485{,}792 & -- \\
    2 & 13{,}878 & 28{,}428 & 6.2 \\
    3 & 1{,}008 & 1{,}617 & 7.3 \\
    \hline
  \end{tabular}
\end{table}

The node ratio is computed with respect to the lower layer. 
The node count decreases steadily toward higher layers while maintaining enough edges to preserve a connected backbone of the environment. 
Across the observed scenes, the higher layers produced coarse connectivity patterns that were visually consistent across multiple traversals of the same areas, even though no explicit loop-closure or topological simplification module was employed. 
A more quantitative assessment of topological quality, for example in terms of path-length distortion, loop and connectivity statistics, or homology-based indicators, is left for future work.

\subsection{Scaling Factor Analysis}
\begin{figure}[tb]
  \centering
  \includegraphics[width=0.9\linewidth]{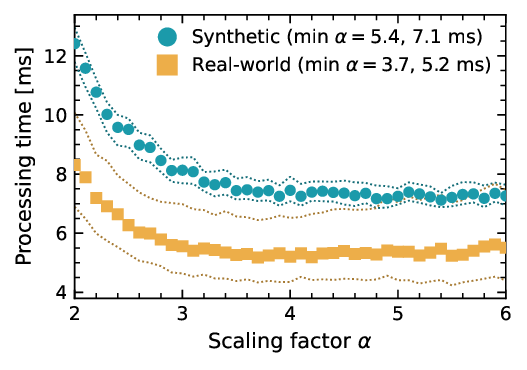}
  \caption{Runtime variation with scaling factor $\alpha$ on synthetic and real-world datasets.}
  \label{fig:alpha_validation}
\end{figure}

To evaluate robustness against parameter changes, the scaling factor $\alpha$ was varied in increments of $0.01$, as shown in Fig.~\ref{fig:alpha_validation}. 
The median processing time reached its minimum at $\alpha=5.4$ for the synthetic dataset and $\alpha=3.7$ for the real-world data. 
In both cases, performance remained nearly constant within the ranges $\alpha \in [3.5, 6.0]$ (synthetic) and $\alpha \in [3.1, 5.7]$ (real-world), which lie close to the favorable region suggested by the complexity analysis in Section~\ref{seq:vigi_para}. 
These results indicate that MLATC does not require fine-tuning of $\alpha$; a fixed setting of $\alpha=4.0$ provides a robust and environment-independent choice within a broad plateau of good performance.

\subsection{Discussion}
The proposed hierarchical topological mapping method achieves real-time operation ($<10$\,ms per frame) across both synthetic and real-world datasets under the tested conditions. 
Its multi-layer representation reduces node redundancy while, in the examined datasets, qualitatively preserving geometric and topological consistency, yielding compact and structurally interpretable maps. 
Compared with the original ATC-DT baseline, MLATC achieves markedly lower computation time and higher throughput, especially in large-scale regimes. 

The theoretical analysis predicts approximately logarithmic search complexity under an idealized uniform planar model, whereas real-world environments exhibit complex and non-uniform spatial distributions, leading to frame-to-frame variations in computational load. 
As a result, the empirical scaling cannot be classified as strictly logarithmic, but it is clearly sublinear and significantly deviates from the linear trend observed for ATC-DT, leading to much shorter processing times over the explored range of map sizes. 
Overall, these results indicate a favorable balance between computational efficiency, scalability, and structural fidelity, suggesting that MLATC is well suited for large-scale topological mapping and long-term spatial reasoning.

\section{Conclusion} \label{sec:conclusion}

This paper presented Multi-Layer ATC (MLATC), a hierarchical extension of an ART-based topological clustering algorithm, designed for fast global topological map building from sequential 3D LiDAR point clouds. 
By recursively using lower-layer nodes as inputs to upper layers and explicitly maintaining parent--child relationships, MLATC builds a multi-scale graph representation that captures both fine local structures and coarse global connectivity within a unified framework. 
Building on the stability--plasticity mechanisms of ART, the proposed architecture preserves the continual-learning capability of the original ATC-DT while addressing its primary limitation, namely the linear growth of the nearest-neighbor search cost with respect to the number of nodes.

We analyzed the vigilance parameter design and search complexity under an idealized uniform planar distribution assumption and showed that, within this model, the expected number of layers grows logarithmically with the total number of nodes and that the size of the candidate set for nearest-neighbor search can be uniformly bounded across layers. 
This analysis led to a practical design rule for the scaling factor, suggesting that a moderate range around $\alpha \approx 3.5\text{--}4.5$ achieves a favorable balance between hierarchy depth and memory usage. 
Although these results rely on simplified assumptions, they provide a useful guideline for parameter selection, and the derived complexity suggests that the overall search cost approaches approximately logarithmic scaling with respect to the number of nodes.

Extensive experiments on synthetic and real-world LiDAR datasets supported these analytical insights. 
In synthetic environments, MLATC transformed the linear scaling of the original ATC-DT into a clearly sublinear, approximately logarithmic trend, achieving millisecond-level per-frame runtimes even when the total number of nodes reached $10^7$. 
On real campus-scale data, the hierarchical map maintained up to six active layers and consistently processed frames within a few milliseconds, remaining well below the $10$\,ms real-time threshold while the number of nodes grew beyond $2 \times 10^5$. 
These results indicate that MLATC enables real-time topological mapping in large-scale environments while preserving coherent hierarchical structures that connect local surfaces to global paths.

Future work will focus on introducing quantitative metrics, such as path-length distortion, loop and connectivity statistics, or homology-based indicators, to rigorously evaluate how well the hierarchy preserves navigationally relevant structure across layers. 
We also plan to exploit the framework's multi-attribute capability by explicitly incorporating surface normals, reflectivity, and traversability into downstream applications such as risk-aware path planning, semantic abstraction, and long-term spatial reasoning. 
Integrating MLATC into complete SLAM and navigation pipelines, including loop closure and multi-robot scenarios, is an important next step toward scalable, structure-aware autonomy in complex real-world environments.

\section*{Acknowledgement} \label{sec:ack}
The authors used an AI system (OpenAI ChatGPT, version GPT-5.1 Thinking) to assist in the refinement of English expression and the clarification of several explanations in this manuscript. All technical content, experimental design, and conclusions were created and verified by the authors.

\newpage

 




\vfill

\end{document}